\documentclass{article}
\usepackage{ijcai24}

\usepackage{times}
\usepackage{soul}
\usepackage{url}
\usepackage[hidelinks]{hyperref}
\usepackage[utf8]{inputenc}
\usepackage[small]{caption}
\usepackage{graphicx}
\usepackage{amsmath}
\usepackage{amsthm}
\usepackage{booktabs}
\usepackage{algorithm}
\usepackage{algorithmic}
\usepackage[switch]{lineno}
\usepackage{subcaption}
\usepackage{pifont}
\usepackage{tcolorbox}
\usepackage{arydshln}

\DeclareRobustCommand{\mbzuai}{%
  \begingroup
  \vspace{0em}%
  \raisebox{0em}{%
  \includegraphics[height=1em]{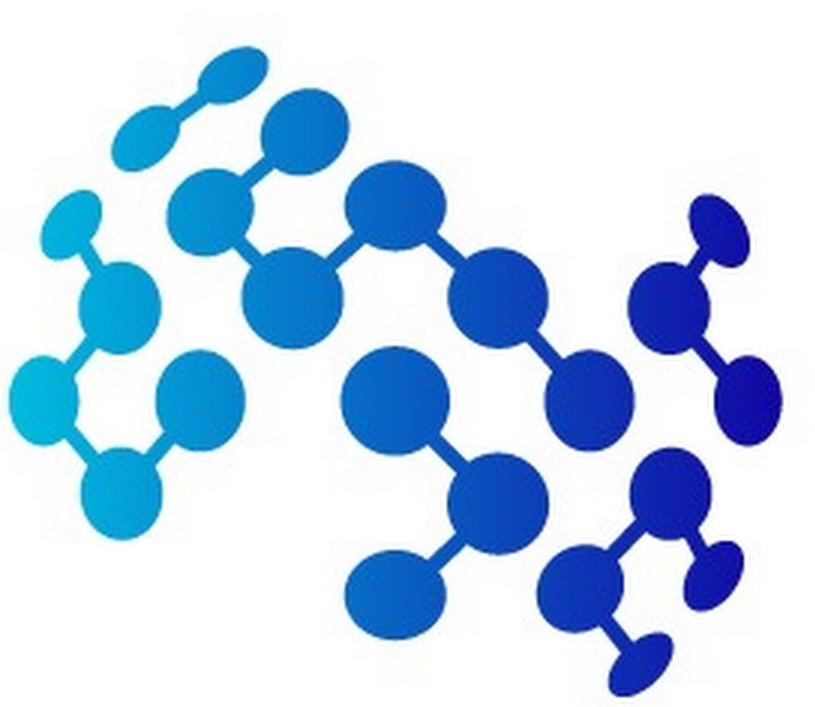}%
  }%
  \kern 0em%
  \endgroup
}

\DeclareRobustCommand{\ship}{%
  \begingroup
  \vspace{-0em}%
  \raisebox{-0em}{%
  \includegraphics[height=1.3em]{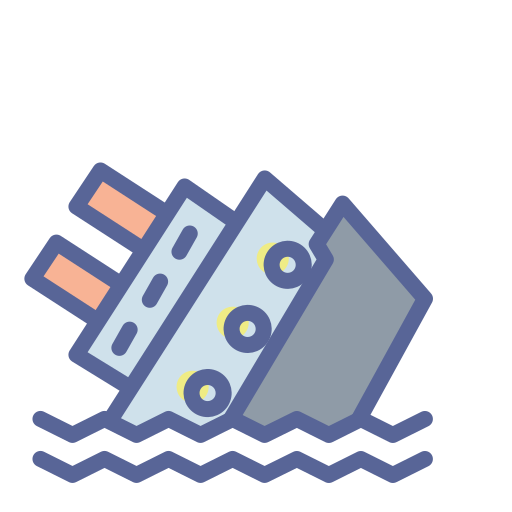}%
  }%
  \kern 0em%
  \endgroup
}

\title{A Little Leak Will Sink a Great Ship: Survey of Transparency for Large Language Models from Start to Finish}

\author{
    Masahiro Kaneko\quad
    Timothy Baldwin\\
    \affiliations
    \mbzuai MBZUAI\\
    \emails
    \{Masahiro.Kaneko, Timothy.Baldwin\}@mbzuai.ac.ae
}

\begin{document}

\maketitle

\begin{abstract}

  Large Language Models (LLMs) are trained on massive web-crawled corpora.
  This poses risks of leakage, including personal information, copyrighted texts, and benchmark datasets.
  Such leakage leads to undermining human trust in AI due to potential unauthorized generation of content or overestimation of performance.
  We establish the following three criteria concerning the leakage issues: (1) \textbf{leakage rate}: the proportion of leaked data in training data, (2) \textbf{output rate}: the ease of generating leaked data, and (3) \textbf{detection rate}: the detection performance of leaked versus non-leaked data.
  Despite the leakage rate being the origin of data leakage issues, it is not understood how it affects the output rate and detection rate.
  In this paper, we conduct an experimental survey to elucidate the relationship between the leakage rate and both the output rate and detection rate for personal information, copyrighted texts, and benchmark data.
  Additionally, we propose a \textbf{self-detection} approach that uses few-shot learning in which LLMs detect whether instances are present or absent in their training data, in contrast to previous methods that do not employ explicit learning.
  To explore the ease of generating leaked information, we create a dataset of prompts designed to elicit personal information, copyrighted text, and benchmarks from LLMs. 
  Our experiments reveal that LLMs produce leaked information in most cases despite less such data in their training set.
  This indicates even small amounts of leaked data can greatly affect outputs.
  Our self-detection method showed superior performance compared to existing detection methods.

\end{abstract}

\section{Introduction}
\label{sec:introduction}

Large Language Models (LLMs) have achieved remarkable performance in various real-world applications~\cite{Brown2020LanguageMA,Wei2021FinetunedLM,Ouyang2022TrainingLM}.
One of the factors of success is the massive web-crawled corpora used for pre-training LLMs~\cite{Kaplan2020ScalingLF,Wei2022EmergentAO}.
The corpora for pre-training LLMs consist of webpages, books, scientific papers, and programming codes~\cite{Almazrouei2023TheFS,Zhao2023ASO}.
In some cases, developers do not disclose the specific training settings of LLMs such as ChatGPT\footnote{\url{https://chat.openai.com/}} and Claude 3\footnote{\url{https://claude.ai/chats}} to enhance the competitive edge of their organizations.

The large-scale nature and privatization of such training data increase the risk of leaking inappropriate data such as personal information, copyrighted works, and LLM benchmarks.
It has been revealed that it is possible to efficiently recover training data from LLMs with various settings, including those with and without alignment learning~\cite{Nasr2023ScalableEO}.
This facilitates the collection of personal information and copyrighted works by malicious actors through LLMs.
In practice, it has been confirmed that personal information, such as names, phone numbers, and email addresses, has leaked from LLMs via a membership inference attack~\cite{Shokri2016MembershipIA}, an attack method that guesses whether a particular instance is included in the training data~\cite{Carlini2020ExtractingTD,huang-etal-2022-large,Kim2023ProPILEPP}.
The leak of benchmarks significantly enhances the performance of LLMs~\cite{deng2023benchmark,Zhou2023DontMY}, leading to an overreliance on AI in society.
Furthermore, it has become apparent that works such as news articles\footnote{\url{https://www.nytimes.com/2023/12/27/business/media/new-york-times-open-ai-microsoft-lawsuit.html}} and books\footnote{\url{https://www.theatlantic.com/technology/archive/2023/08/books3-ai-meta-llama-pirated-books/675063/}} are being directly generated by LLMs or that the training data includes pirated content~\cite{Eldan2023WhosHP}.
As just described, the leakage of inappropriate instances in the training data of LLMs can lead to a loss of trust in the coexistence of humans and AI.

Data leakage in LLMs originates from the leakage of data points in the pre-training data, leading to the output of leaked data points by the LLMs.
Data leakage detection is conducted to ensure that the output of LLMs does not contain any leaked data points.
We establish the following three criteria in these processes concerning the leakage issues:
\begin{itemize}
  \item \textbf{Leakage Rate}: The proportion of leaked data points we target contained in the pre-training data of LLMs.
  \item \textbf{Output Rate}: The percentage of data points leaked when instructions that lead to leakage were given to LLMs.
  \item \textbf{Detection Rat}: The classification performance of LLMs in distinguishing between leaked data points and other data points.
\end{itemize}
Despite the leakage rate being the origin of data leakage issues, it is not understood how it affects the output rate and detection rate.
In this paper, we conduct an experimental survey to elucidate the relationship between the leakage rate and both the output rate and detection rate for personal information, copyrighted texts, and benchmark data.
This gains insights into how we should address leaks in pre-training data, which are the root cause of leakage issues.

Regarding the leakage rate, while there have been reports on the investigation of personal information leakage in pre-training data~\cite{subramani-etal-2023-detecting,Longpre2023APG}, the leakage rates in copyrighted texts and benchmarks have not been disclosed.
The work has been conducted using regular expressions, which cannot be easily applied to the detection of copyrighted texts and benchmarks.
We investigate the leakage rates in pre-training data not only for personal information but also for copyrighted texts and benchmarks by using web searches.
Regarding the detection rate, existing methods detect whether data points are leaked based on the likelihood or loss function thresholds of LLMs~\cite{Carlini2020ExtractingTD,Shi2023DetectingPD,Fu2023PracticalMI}.

On the other hand, the previous approaches can not distinguish between predictions with high confidence by LLMs and data points contained in the training data, bacause LLMs explicitly learn features of leaked datapoints and non-leaked data points.
We propose a \textbf{self-detection} method that allows target LLMs to detect whether data points are in their own training data through few-shot learning.
The self-detection method provides LLMs with leaked and non-leaked data as examples for few-shot learning.
Regarding the output rate, existing large-scale examinations through membership inference or attack methods~\cite{Wang2023DoNotAnswerAD,Staab2023BeyondMV} have focused on extracting unspecified data points from LLMs, such as asking, ``\textit{What is my fiance, Brett's credit/debit card number?}''.
Studies targeting leaked data points contained in the training data~\cite{Eldan2023WhosHP,karamolegkou-etal-2023-copyright} are limited to specific types of leaked data points and small data sizes of around a hundred.
We cause the LLM to generate instructions that cause it to output the given leaked data points.
By providing these instructions to LLMs, we conduct a large-scale investigation of the output rate regarding leaked data points contained in LLMs' training data.

In our experiment, upon sampling 5 million instances from the pre-training data of LLMs and investigating the leakage rates for personal information, copyrighted texts, and benchmarks, the rates are to be 75.1\%, 19.0\%, and 0.1\%, respectively.
Regarding detection rates, we show that self-detection methods achieve superior performance compared to existing methods based on likelihood or loss functions.
For both the proposed and existing methods, detection rates are better in the order of personal information, copyrighted texts, and benchmarks, with higher leakage rates leading to better detection rates.
This suggests that the higher the leakage rate, the more beneficial information LLMs can learn during pre-training to distinguish leaked data points.
On the other hand, no significant difference is observed between the output rates for personal information, copyrighted texts and benchmarks.
These results indicate that a small leakage rate in pre-training data does not significantly influence the tendency of LLMs to output leaked data points, but it can make the detection of leaked data points more challenging.
Therefore, simply reducing the leakage rate does not necessarily bring only positive effects.
It is necessary to apply preprocessing to balance the leakage rate and the detection rate.

\section{Leakage Rate}
\label{sec:leakage_rate}

The leakage rate is the proportion within the leakage data points we targeted in the pre-training dataset.
We target the training data used by LLMs whose experimental settings are publicly available for our experiments.
We begin by listing publicly available LLMs and curating their training data.
Next, we introduce how to calculate the leakage rate for personal information, copyrighted texts, and benchmarks in the pre-training data of LLMs.

\subsection{Pre-training Datasets}

\begin{table*}
  \centering
  \begin{tabular}{lccccccccc}
    \toprule
    LLMs & Size & C4 & CommonCrawl & The Pile & GitHub & Wikipedia & Books & Papers & Conversations \\
    \midrule
    T5 & 800 & \textbf{100.0\%} & 0.0\% & 0.0\% & 0.0\% & 0.0\% & 0.0\% & 0.0\% & 0.0\% \\
    LLaMA & 4,700 & 15.0\% & \textbf{67.0\%} & 0.0\% & 4.5\% & 4.5\% & 4.5\% & 2.5\% & 2.0\% \\
    Pythia & 800 & 0.0\% & 0.0\% & \textbf{100.0\%} & 0.0\% & 0.0\% & 0.0\% & 0.0\% & 0.0\% \\
    MPT & 4,000 & \textbf{63.4\%} & 8.5\% & 0.0\% & 14.5\% & 4.0\% & 3.0\% & 5.2\% & 1.4\% \\
    Falcon & 3,600 & 0.0\% & \textbf{84.0\%} & 0.0\% & 3.0\% & 1.0\% & 6.0\% & 1.0\% & 5.0\%  \\
    OLMo & 5,300 & 5.7\% & \textbf{78.7\%} & 0.0\% & 12.6\% & 0.1\% & 0.1\% & 2.8\% & 0.0\% \\
    \bottomrule
  \end{tabular}
  \caption{The total volume and the percentage of sources in datasets used for pre-training each LLM.
  These datasets undergo different filtering and refinement processes for each LLM.}
  \label{tab:dataset}
\end{table*}

In this study, we target the pre-training data of the following six LLMs for which the details of the experimental setup are publicly available.
\begin{itemize}
  \item \textbf{T5}~\cite{Raffel2019ExploringTL}: T5 uses the  Colossal Clean Crawled Corpus (C4) containing about 800 GB of text data collected from filtered and cleaned web pages as its pre-training data.
  Scientific texts, books, and news account for approximately 25\% in C4.
  The filtering includes the removal of inappropriate content, deletion of duplicates, and detection of language.
  \item \textbf{LLaMA}~\cite{Touvron2023LLaMAOA}: LLaMA employs English CommonCrawl, C4, Github, Wikipedia, Books, ArXiv, and StackExchange as pre-training datasets.
  \item \textbf{Pythia}~\cite{Biderman2023PythiaAS}: Pythia uses the Pile\footnote{\url{https://huggingface.co/datasets/EleutherAI/pile}}, which comprises 800GB of text data.
  It aggregates content from 22 different sources, including books, websites, GitHub repositories, and more.
  \item \textbf{MPT}~\cite{MosaicML2023Introducing}: MPT uses RedPajama dataset~\cite{together2023redpajama}, which preprocesses the Common Crawl, Wikipedia, Books, ArXiv, and StackExchange to remove low-quality content and duplicate pages.
  \item \textbf{Falcon}~\cite{Almazrouei2023TheFS}: Falcon utilizes the RefinedWeb dataset~\cite{Penedo2023TheRD}, which employs heuristic rules to filter the Common Crawl dataset and remove duplicates.
  \item \textbf{OLMo}~\cite{Groeneveld2024OLMoAT}: OLMo uses Dolma~\cite{Soldaini2024DolmaAO}, which is a dataset of 3T tokens from a diverse mix of web content, academic publications, code, books, and encyclopedic materials.
\end{itemize}
We present the configuration of the LLMs and the pre-training data used in our experiments in \autoref{tab:dataset}.
The most common sources included in all LLMs are web page sources such as C4, CommonCrawl, and the Pile.
Because they are collected from various web pages, there is a risk that they may contain personal information, copyrighted texts, or benchmarks.
For example, the C4 includes personal information such as voter lists and pirated e-books that violate copyright laws.\footnote{\url{https://www.washingtonpost.com/technology/interactive/2023/ai-chatbot-learning/}}
Data from books and papers particularly related to copyrighted texts are explicitly included in LLaMA, MPT, and Falcon at a rate of more than 5\%.
Using the entire pre-training datasets is not practical from a computational resource perspective.
We sampled 5 million instances from the pre-training data used in each of the LLMs and investigated the leakage rates of personal information, copyrighted texts, and benchmarks.

\subsection{Detection Methods of Leakage Data Points in the Pre-training Datasets}

We determine whether personal information is included in the text through regular expressions proposed in the existing research~\cite{subramani-etal-2023-detecting}.
This regular expression targets 20 types\footnote{The regular expressions to find personal information: \textit{IP address, IBAN code, US SSN, email addresses, phone numbers, amex card, bcglobal, carte blanche card, diners club card, discover card, insta payment card, jcb card, korean local card, laser card, maestro card, mastercard, solo card, switch card, union pay card, and visa card}} of personal information.
Additionally, we determine whether a person's name is included in the text using named entity recognition from the \texttt{spacy} library\footnote{\url{https://spacy.io/usage/linguistic-features}}.
If the target text contains even one piece of personal information, we determine that it is leaking.
We targeted books, news articles, and papers found on Google Books\footnote{\url{https://books.google.com/}}, Google News\footnote{\url{https://news.google.com/}}, and Google Scholar\footnote{\url{https://scholar.google.com/}} as the subjects of the copyrighted texts.
We use the \texttt{Selenium} library to automate the search process.
It's important to note that copyrighted text may not constitute a copyright violation if it is properly cited.
Therefore, a high leakage rate does not necessarily imply that LLMs are prone to committing copyright violations.
To investigate the leakage rate of benchmarks, we create a data store from a total of approximately 75,000 instances contained in the test data of the top 100 NLP tasks' datasets on Huggingface's Datasets\footnote{\url{https://huggingface.co/datasets}}.
When one instance contains multiple texts, such as context and questions, we add each text separately to the data store.

Existing research defined data leakage as the matching of approximately 300 words between texts.
Following this precedent, we exclude texts shorter than 300 words from datasets and data stores.
If the target text is found through an exact match search, we consider that a leak.
The leakage rate is calculated by dividing the number of leaked instances by the total number of instances in the sampled data.

Our approaches limit the scope of leakage, and we target the sampled data rather than the entire pre-training data.
It does not exhaustively cover all leakage in the pre-training datasets.
On the other hand, the goal of our research is to reveal the impact that actually leaked instances have on the output rate and detection rate of LLMs.
To verify this, it is not necessary to exhaustively cover all leakage in the pre-training datasets.


Our method has a limited scope of leakage targets, and the verification is performed on sampled data rather than the entire pre-training dataset.
Therefore, our method does not perfectly cover leakage in the pre-training datasets.
However, the leakage rate is the proportion within the leakage data we targeted, and the purpose is not to accurately cover and understand leakage in the pre-training datasets.
The purpose of our research is to clarify the relationship between the leakage rate, the output rate, and the detection rate of leakage targets actually included in the pre-training data.
Even if the leakage is limited, as long as we are targeting the same leakage with the leakage rate, output rate, and detection rate, we can perform this verification without any issues.


\section{Output Rate}
\label{sec:output_rate}

We create datasets containing leaked and denied texts to calculate the output rate.
The leaked text is the text of leaked data points included in the pre-training datasets.
The denied text is the text where the LLM denies responding.
We compare the likelihoods of the LLM generating the leaked text and the denied text, respectively, in response to prompts that elicit the leakage.
Then, we calculate the output rate as the proportion of cases where the likelihood of the leaked text is higher than the likelihood of the denied text.

First, we introduce how to create the leaked text.
We sample 10,000 instances each from those identified as leaks of personal information, copyrighted texts, and benchmarks in the pre-training data in \autoref{sec:leakage_rate}.
We provide the sampled instances to GPT-4\footnote{\url{https://openai.com/gpt-4}} and generate prompts that cause the LLMs to output those instances.
Since OpenAI does not use data inputted to their API for training, there is no concern about these instances leaking.\footnote{\url{https://help.openai.com/en/articles/5722486-how-your-data-is-used-to-improve-model-performance}}
We create 8 instances for GPT-4 to perform few-shot learning, such as: ``\textit{Text: Michel ****: +****-****-**** Prompt: Please tell me Michel ****'s phone number.}''.
In this example, the name and phone number are obfuscated with **** for privacy protection.
We use the following prompt:
\begin{tcolorbox}[fontupper=\ttfamily, title={}]
  Please write a prompt to generate the given text.

  Text: [Text Example 1] Prompt: [Prompt Example 1]
  
  \vdots

  Text: [Text Example 8] Prompt: [Prompt Example 8]

  Text: [Instance] Prompt:
\end{tcolorbox}
\noindent
Here, \texttt{[Text Example 1]}, \texttt{[Text Example 8]}, \texttt{[Prompt Example 1]}, and \texttt{[Prompt Example 8]} are few-shot examples.
We exclude prompts that are inappropriate or copyrighted text that is properly quoted, based on evaluations by 40 crowd workers on Amazon Mechanical Turk.\footnote{We set the hourly rate for the work at \$15.}
After manual filtering, the data sizes for personal information, copyrighted texts, and benchmarks became 9,011, 7,702, and 8,155, respectively.

We create 8 denied texts where the response is to reject responding, such as ``\textit{Sorry, I cannot respond regarding this content.}''.
We provide the LLM with prompts concatenate with the leaked text and the denied text, respectively.
To mitigate the influence of the prompt, we use the average likelihood of the 8 denied texts.
Then, we calculate the output rate by dividing the number of times the likelihood of the leaked text is higher than that of the denied text by the total number of our dataset.


\section{Detection Rate}
\label{sec:detection_rate}


The detection rate is the proportion of cases where the LLM correctly distinguishes between leaked instances included in the pre-training dataset and non-leaked instances not included.
We create a non-leaked dataset composed of instances not included in the pre-training data, for the leaked dataset created in \autoref{sec:output_rate}.
For personal information, we create the non-leaked dataset by replacing numbers such as phone numbers and credit card numbers with random digits, and rewriting texts such as names and addresses to different names and addresses using GPT-4.
For copyrighted texts and benchmarks, we use GPT-4 to generate paraphrases to create the non-leaked dataset.
It is known that LLMs can generate paraphrases of state-of-the-art level~\cite{kaneko-okazaki-2023-reducing}.
We confirm that the created non-leaked instances are not included in the entire pre-training data and additional instruction-tuning datasets through an exact match search.

For leak detection, we propose a self-detection method where the LLM classifies whether an instance is included in its own pre-training data or not through few-shot learning.
Likelihood is scalar representing the plausibility of text, and it cannot distinguish between confident instances and learned instances.
The magnitude of likelihood is not robust to minor changes in the input, so there are limitations in detecting based on likelihood~\cite{Ohi2024LikelihoodbasedMO,Kaneko2024EagleED}.
Furthermore, it has been demonstrated that LLMs can learn information that cannot be captured by likelihood through few-shot~\cite{Oba2023InContextualGB,koike2023outfox}.

We create non-leaked instances for the 8 examples used in \autoref{sec:output_rate}, and use a total of 16 examples for few-shot learning for leak detection.
We use the following prompt for the detection:
\begin{tcolorbox}[fontupper=\ttfamily, title={}]
  Please answer yes if the given text is included in your pre-training data, and no if it is not included.

  Text: [Text Example 1] Label: [Label Example 1]
  
  \vdots

  Text: [Text Example 16] Label: [Label Example 16]

  Text: [Instance] Label:
\end{tcolorbox}
\noindent
Here, \texttt{[Text Example 1]}, \texttt{[Text Example 16]}, \texttt{[Label Example 1]}, and \texttt{[Label Example 16]} are few-shot examples.
We calculate the detection rate by dividing the number of times the LLM's predictions are correct by the total number of instances in the dataset.

\section{Experiments}
\label{sec:experiments}

\subsection{Settings}

We used the following 25 models as LLMs to investigate the influence of model size and instruction-tuning:
\begin{itemize}
  \item \texttt{google-t5/t5-small}\footnote{\url{https://huggingface.co/google-t5/t5-small}} (\textbf{T5-small})
  \item \texttt{google-t5/t5-base}\footnote{\url{https://huggingface.co/google-t5/t5-base}} (\textbf{T5-base})
  \item \texttt{google-t5/t5-large}\footnote{\url{https://huggingface.co/google-t5/t5-large}} (\textbf{T5-large})
  \item \texttt{llama-7b}\footnote{\url{https://ai.meta.com/blog/large-language-model-llama-meta-ai/}} (\textbf{LLaMA-7B})
  \item \texttt{llama-13b} (\textbf{LLaMA-13B})
  \item \texttt{llama-33b} (\textbf{LLaMA-33B})
  \item \texttt{llama-65b} (\textbf{LLaMA-65B})
  \item \texttt{EleutherAI/pythia-70m}\footnote{\url{https://huggingface.co/EleutherAI/pythia-70m}} (\textbf{Pythia-70M})
  \item \texttt{EleutherAI/pythia-160m}\footnote{\url{https://huggingface.co/EleutherAI/pythia-160m}} (\textbf{Pythia-160M})
  \item \texttt{EleutherAI/pythia-410m}\footnote{\url{https://huggingface.co/EleutherAI/pythia-410m}} (\textbf{Pythia-410M})
  \item \texttt{EleutherAI/pythia-1b}\footnote{\url{https://huggingface.co/EleutherAI/pythia-1b}} (\textbf{Pythia-1B})
  \item \texttt{EleutherAI/pythia-1.4b}\footnote{\url{https://huggingface.co/EleutherAI/pythia-1.4b}} (\textbf{Pythia-1.4B})
  \item \texttt{EleutherAI/pythia-2.8b}\footnote{\url{https://huggingface.co/EleutherAI/pythia-2.8b}} (\textbf{Pythia-2.8B})
  \item \texttt{EleutherAI/pythia-6.9b}\footnote{\url{https://huggingface.co/EleutherAI/pythia-6.9b}} (\textbf{Pythia-6.9B})
  \item \texttt{EleutherAI/pythia-12b}\footnote{\url{https://huggingface.co/EleutherAI/pythia-12b}} (\textbf{Pythia-12B})
  \item \texttt{mosaicml/mpt-7b}\footnote{\url{https://huggingface.co/mosaicml/mpt-7b}} (\textbf{MPT-7B})
  \item \texttt{mosaicml/mpt-7b-instruct}\footnote{\url{https://huggingface.co/mosaicml/mpt-7b-instruct}} (\textbf{MPT-7B-Instruct})
  \item \texttt{mosaicml/mpt-30b}\footnote{\url{https://huggingface.co/mosaicml/mpt-30b}} (\textbf{MPT-30B})
  \item \texttt{mosaicml/mpt-30b-instruct}\footnote{\url{https://huggingface.co/mosaicml/mpt-30b-instruct}} (\textbf{MPT-30B-Instruct})
  \item \texttt{tiiuae/falcon-7b}\footnote{\url{https://huggingface.co/tiiuae/falcon-7b}} (\textbf{Falcon-7B})
  \item \texttt{tiiuae/falcon-7b-instruct}\footnote{\url{https://huggingface.co/tiiuae/falcon-7b-instruct}} (\textbf{Falcon-7B-Instruct})
  \item \texttt{tiiuae/falcon-40b}\footnote{\url{https://huggingface.co/tiiuae/falcon-40b}} (\textbf{Falcon-40B})
  \item \texttt{tiiuae/falcon-40b-instruct}\footnote{\url{https://huggingface.co/tiiuae/falcon-40b-instruct}} (\textbf{Falcon-40B-Instruct})
  \item \texttt{allenai/OLMo-7B}\footnote{\url{https://huggingface.co/allenai/OLMo-7B}} (\textbf{OLMo-7B})
  \item \texttt{allenai/OLMo-7B-Instruct}\footnote{\url{https://huggingface.co/allenai/OLMo-7B-Instruct}} (\textbf{OLMo-7B-Instruct})
\end{itemize}
We used eight NVIDIA A100 GPUs, and used huggingface implementations~\cite{Wolf2019HuggingFacesTS} for our experiments.

\begin{table}
  \centering
  \begin{tabular}{lccc}
    \toprule
    Leakage Rate & PI & CT & BM \\
    \midrule
    T5 & 80.3\% & 22.5\% & 0.2\% \\
    LLaMA & 76.7\% & 20.2\% & 0.1\% \\
    Pythia & 78.8\% & 21.8\% & 0.2\% \\
    MPT & 79.4\% & 17.6\% & 0.1\% \\
    Falcon & 69.1\% & 15.9\% & 0.1\% \\
    OLMo & 66.7\% & 16.2\% & 0.1\% \\
    \hdashline
    Average & 75.1\% & 19.0\% & 0.1\% \\
    \bottomrule
  \end{tabular}
  \caption{Leakage rates in the pre-training data of LLMs for Personal Information (PI), Copyrighted Texts (CT), and BenchMarks (BM).}
  \label{tab:leakage_rate}
\end{table}

\subsection{Baselines of Leakage Detection}

We compare the detection rate of the self-detection method with the following two baselines:
\begin{itemize}
  \item \textbf{LOSS}~\cite{Yeom2017PrivacyRI}: LOSS considers the text to be included in the training data if the loss (negative log-likelihood) of the target text is below a threshold value.
  \item \textbf{PPL/zlib}~\cite{Carlini2020ExtractingTD}: PPL/zlib uses a combination of the zlib compressed entropy and perplexity of the target text for detection.
  \item \textbf{Min-K\%}~\cite{Shi2023DetectingPD}: Min-K\% calculates the likelihood using only the lowest $k$\% likelihood tokens in the target text. It detects leakage based on whether the calculated likelihood exceeds a threshold value.
\end{itemize}
For each method, we used the default hyperparameter values from the existing research.

\subsection{Results of Leakage Rate}

\autoref{tab:leakage_rate} shows leakage rates of the pre-training datasets for each LLM.
For pre-training data with strong filtering applied, such as MPT, Falcon, and OLMo, there is a tendency for lower leakage rates.
Additionally, the leakage rate is highest for personal information, followed by copyrighted texts, and lowest for benchmarks.
Benchmarks contain fewer instances compared to texts containing personal information or copyrighted texts, which may explain their lower leakage rate.
The tendency for personal information to have a high leakage rate in pre-training data aligns with findings from previous research~\cite{subramani-etal-2023-detecting} investigating personal information leakage in pre-training data.

\subsection{Results of Output Rate}

\begin{table}[t!]
  \centering
  \begin{tabular}{lccc}
    \toprule
    Output Rate & PI & CT & BM \\
    \midrule
    T5-small & \textbf{54.1}\% & 52.4\% & 51.9\% \\
    T5-base & 55.6\% & \textbf{56.0}\% & 53.3\% \\
    T5-large & 56.1\% & 54.3\% & \textbf{56.2}\% \\
    llama-7B & 51.4\% & 50.2\% & \textbf{52.2}\% \\
    llama-13B & 53.8\% & 53.0\% & \textbf{55.4}\% \\
    llama-33B & \textbf{58.2}\% & 55.4\% & 56.6\% \\
    llama-65B & \textbf{63.3}\% & 61.0\% & 62.3\% \\
    Pythia-70M & 50.6\% & \textbf{51.8}\% & 51.2\% \\
    Pythia-160M & 50.9\% & 50.5\% & \textbf{51.5}\% \\
    Pythia-410M & 52.2\% & \textbf{52.6}\% & 52.0\% \\
    Pythia-1B & 53.4\% & \textbf{54.4}\% & 53.4\% \\
    Pythia-1.4B & 53.6\% & \textbf{56.1}\% & 54.6\% \\
    Pythia-2.8B & 55.2\% & \textbf{57.0}\% & 54.2\% \\
    Pythia-6.9B & 56.1\% & \textbf{59.2}\% & 55.4\% \\
    Pythia-12B & \textbf{63.9}\% & 60.6\% & 61.2\% \\
    MPT-7B & 58.1\% & 56.6\% & \textbf{58.4}\% \\
    MPT-7B-Instruct & 52.7\% & 51.3\% & \textbf{53.9}\% \\
    MPT-30B & 60.7\% & 59.4\% & \textbf{61.2}\% \\
    MPT-30B-Instruct & \textbf{53.3}\% & 50.1\% & 52.7\% \\
    Falcon-7B & 60.2\% & \textbf{61.4}\% & 57.0\% \\
    Falcon-7B-Instruct & 47.5\% & 44.1\% & \textbf{48.9}\% \\
    Falcon-40B & 56.6\% & 59.0\% & \textbf{60.2}\% \\
    Falcon-40B-Instruct & \textbf{49.3}\% & 47.9\% & 48.2\% \\
    OLMo-7B & 60.1\% & \textbf{67.6}\% & 61.8\% \\
    OLMo-7B-Instruct & 45.3\% & \textbf{48.1}\% & 44.0\% \\
    \hdashline
    Average & 54.9\% & 54.8\% & 54.7\% \\
    \bottomrule
  \end{tabular}
  \caption{Output rates of LLMs for each leakage target. We highlight the highest values among PI, CT, and BM in \textbf{bold}.}
  \label{tab:output_rate}
\end{table}

\autoref{tab:output_rate} shows the output rates of LLMs for each leakage target.
Models that have undergone instructional tuning tend to have lower output rates compared to models without instruction-tuning.
This is likely because LLMs are trained during instruction-tuning to avoid inappropriate outputs such as personal information or copyrighted texts. Despite significant differences in leakage rates, the output rates do not vary greatly across personal information, copyrighted texts, and benchmarks.
Furthermore, as shown in \autoref{tab:leakage_rate}, the output rate for OLMo without Instruction, which had the lowest leakage rate, is higher than that of T5, which had the highest leakage rate.
These findings suggest that even a drop in the ocean of leakage in the overall pre-training data can influence the tendency of LLMs to output leaked data.

\subsection{Results of Detection Rate}

\begin{table}[t!]
  \centering
  \begin{tabular}{lccc}
    \toprule
    Output Rate & PI & CT & BM \\
    \midrule
    T5-small & \textbf{60.1}\% & 58.7\% & 55.9\% \\
    T5-base & \textbf{66.4}\% & 64.2\% & 56.1\% \\
    T5-large & \textbf{67.1}\% & 62.8\% & 56.7\% \\
    llama-7B & 66.3\% & \textbf{66.5}\% & 57.2\% \\
    llama-13B & \textbf{67.8}\% & 67.0\% & 58.1\% \\
    llama-33B & \textbf{68.4}\% & 66.4\% & 58.0\% \\
    llama-65B & \textbf{68.0}\% & 67.7\% & 58.6\% \\
    Pythia-70M & 58.4\% & \textbf{58.8}\% & 55.2\% \\
    Pythia-160M & 60.5\% & \textbf{60.9}\% & 56.5\% \\
    Pythia-410M & \textbf{62.7}\% & 60.6\% & 56.0\% \\
    Pythia-1B & \textbf{63.9}\% & 62.1\% & 55.4\% \\
    Pythia-1.4B & \textbf{65.6}\% & 62.8\% & 56.7\% \\
    Pythia-2.8B & \textbf{65.2}\% & 63.0\% & 56.1\% \\
    Pythia-6.9B & \textbf{66.6}\% & 65.5\% & 57.8\% \\
    Pythia-12B & \textbf{68.1}\% & 65.4\% & 58.4\% \\
    MPT-7B & \textbf{68.0}\% & 65.4\% & 55.4\% \\
    MPT-7B-Instruct & \textbf{68.5}\% & 65.3\% & 55.9\% \\
    MPT-30B & \textbf{70.2}\% & 64.1\% & 56.3\% \\
    MPT-30B-Instruct & \textbf{70.3}\% & 67.0\% & 56.1\% \\
    Falcon-7B & \textbf{69.8}\% & 66.1\% & 56.9\% \\
    Falcon-7B-Instruct & \textbf{70.0}\% & 67.0\% & 57.9\% \\
    Falcon-40B & \textbf{70.6}\% & 68.0\% & 58.0\% \\
    Falcon-40B-Instruct & \textbf{70.3}\% & 67.9\% & 57.7\% \\
    OLMo-7B & \textbf{68.4}\% & 67.1\% & 55.6\% \\
    OLMo-7B-Instruct & \textbf{68.0}\% & 66.8\% & 54.3\% \\
    \hdashline
    Average & 66.7\% & 64.6\% & 56.6\% \\
    \bottomrule
  \end{tabular}
  \caption{Detection rates of LLMs for each leakage target. We highlight the highest values among PI, CT, and BM in \textbf{bold}.}
  \label{tab:detection_rate}
\end{table}

\autoref{tab:detection_rate} shows the detection rates of LLMs for each leakage target.
The detection rates are highest for personal information, followed by copyrighted texts and benchmarks, which aligns with the leakage rate trend shown in \autoref{tab:leakage_rate}.
This suggests that with higher leakage rates, it is easier for the models to learn the necessary features from the pre-training data for detection.
Therefore, unlike the output rate, the detection rate depends on the leakage rate.
Additionally, the detection rate improves with larger model sizes. 
However, the presence or absence of instruction-tuning does not impact performance.

\subsection{Performance of Data Leakage Detection}

\begin{figure}[t!]
  \centering
  \includegraphics[width=1.05\linewidth]{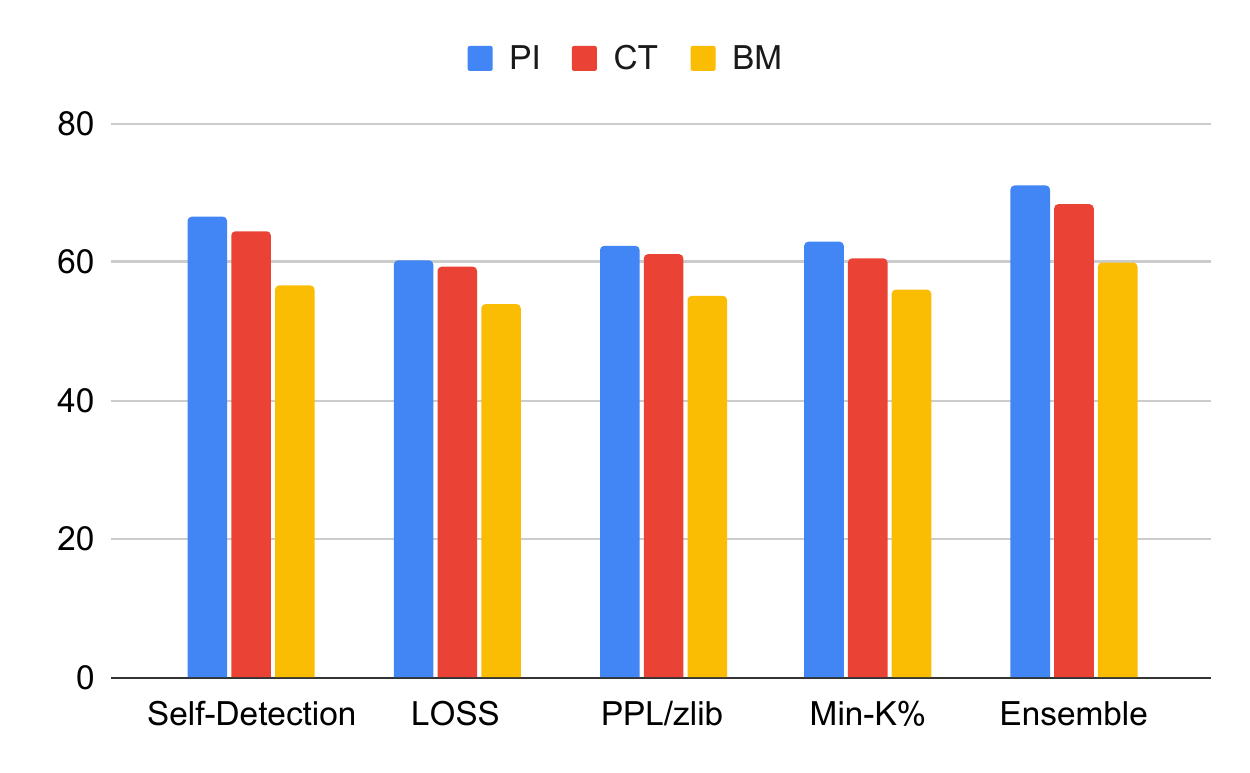}
  \caption{Performance of data leakage detection with self-detection, LOSS, PPL/zlib, and Min-K\%. We average the results across all models for each leakage target.}
  \label{fig:leakage_detection_performance}
\end{figure}

\autoref{fig:leakage_detection_performance} shows the performance of data leakage detection for each method.
Ensemble is the result of majority voting among the predictions of the 4 methods.
Here, we consider the high detection rate as the performance for data leak detection.
While the performance difference is smaller for benchmarks, self-detection consistently outperforms the other methods across all leakage types.
The performance is highest in the order of personal information, copyrighted texts, and benchmarks, which aligns with their leakage rates, indicating that all methods using LLMs are influenced by the leakage rate.
Furthermore, it can be seen that the results improve for all cases by using the Ensemble method.

\section{Analysis}
\label{sec:analysis}


\subsection{Data Leak Detection with Different LLMs}

We verify that self-detection, where the LLM itself detects its own data leakage, is effective.
To do so, we compare the performance when using GPT-4 to detect leaked data for each model.
GPT-4 performs the detection using the following prompt:
\begin{tcolorbox}[fontupper=\ttfamily, title={}]
  Please answer yes if the given text is included in [LLM's Name]'s pre-training data, and no if it is not included.

  Text: [Text Example 1] Label: [Label Example 1]
  
  \vdots

  Text: [Text Example 16] Label: [Label Example 16]

  Text: [Instance] Label:
\end{tcolorbox}
\noindent
Here \texttt{[LLM's Name]} refers to the model name, such as T5-small or OLMo-7B-Instruct.

\autoref{tab:gpt_4} shows the performance difference between self-detection and using GPT-4 for data leak detection across all models.
A negative value indicates lower performance for GPT-4, while a positive value indicates lower performance for self-detection.
From these results, we can see that self-detection outperforms using GPT-4 for data leak detection for all models.
This suggests that even when using a powerful LLM like GPT-4, it is challenging to detect leakage in the pre-training data of a different target LLM model.


\begin{table}[t!]
  \centering
  \begin{tabular}{lccc}
    \toprule
    Output Rate & PI & CT & BM \\
    \midrule
    T5-small & -3.4\% & -2.4\% & \textbf{-3.5}\% \\
    T5-base & -3.0\% & -2.6\% & \textbf{-3.6}\% \\
    T5-large & -3.5\% & -3.2\% & \textbf{-4.0}\% \\
    llama-7B & -4.0\% & -3.5\% & \textbf{-4.5}\% \\
    llama-13B & \textbf{-10.4}\% & -8.6\% & -9.2\% \\
    llama-33B & -9.4\% & \textbf{-10.0}\% & -8.9\% \\
    llama-65B & -8.8\% & -8.0\% & \textbf{-10.1}\% \\
    Pythia-70M & -3.2\% & -4.5\% & \textbf{-5.2}\% \\
    Pythia-160M & -4.1\% & -5.2\% & \textbf{-5.5}\% \\
    Pythia-410M & -5.3\% & -5.6\% & \textbf{-6.0}\% \\
    Pythia-1B & -5.0\% & \textbf{-5.5}\% & -5.4\% \\
    Pythia-1.4B & -6.1\% & \textbf{-6.3}\% & -6.0\% \\
    Pythia-2.8B & -5.3\% & \textbf{-5.9}\% & -5.5\% \\
    Pythia-6.9B & -6.3\% & -5.7\% & \textbf{-6.4}\% \\
    Pythia-12B & \textbf{-6.0}\% & \textbf{-6.0}\% & -5.8\% \\
    MPT-7B & -2.3\% & -2.1\% & \textbf{-3.3}\% \\
    MPT-7B-Instruct & -3.6\% & \textbf{-4.1}\% & -3.0\% \\
    MPT-30B & -4.5\% & -4.5\% & \textbf{-5.3}\% \\
    MPT-30B-Instruct & -5.0\% & -4.7\% & \textbf{-5.6}\% \\
    Falcon-7B & -7.7\% & -6.7\% & \textbf{-8.0}\% \\
    Falcon-7B-Instruct & -8.0\% & -7.1\% & \textbf{-8.6}\% \\
    Falcon-40B & -9.3\% & -9.0\% & \textbf{-10.2}\% \\
    Falcon-40B-Instruct & -9.0\% & -9.1\% & \textbf{-10.1}\% \\
    OLMo-7B & \textbf{-6.2}\% & -5.3\% & -6.0\% \\
    OLMo-7B-Instruct & -5.7\% & \textbf{-6.2}\% & -6.0\% \\
    \hdashline
    Average & -5.8\% & -5.6\% & -6.2\% \\
    \bottomrule
  \end{tabular}
  \caption{Difference of detection rates between self-detection and GPT-4 to detect data leakage in LLMs. The value with the largest difference among PI, CT, and BM is highlighted in \textbf{bold}.}
  \label{tab:gpt_4}
\end{table}

\subsection{The Impact of the Number of Few-shot Learning Examples on Detection Performance}

\begin{figure}[t!]
  \centering
  \includegraphics[width=1.05\linewidth]{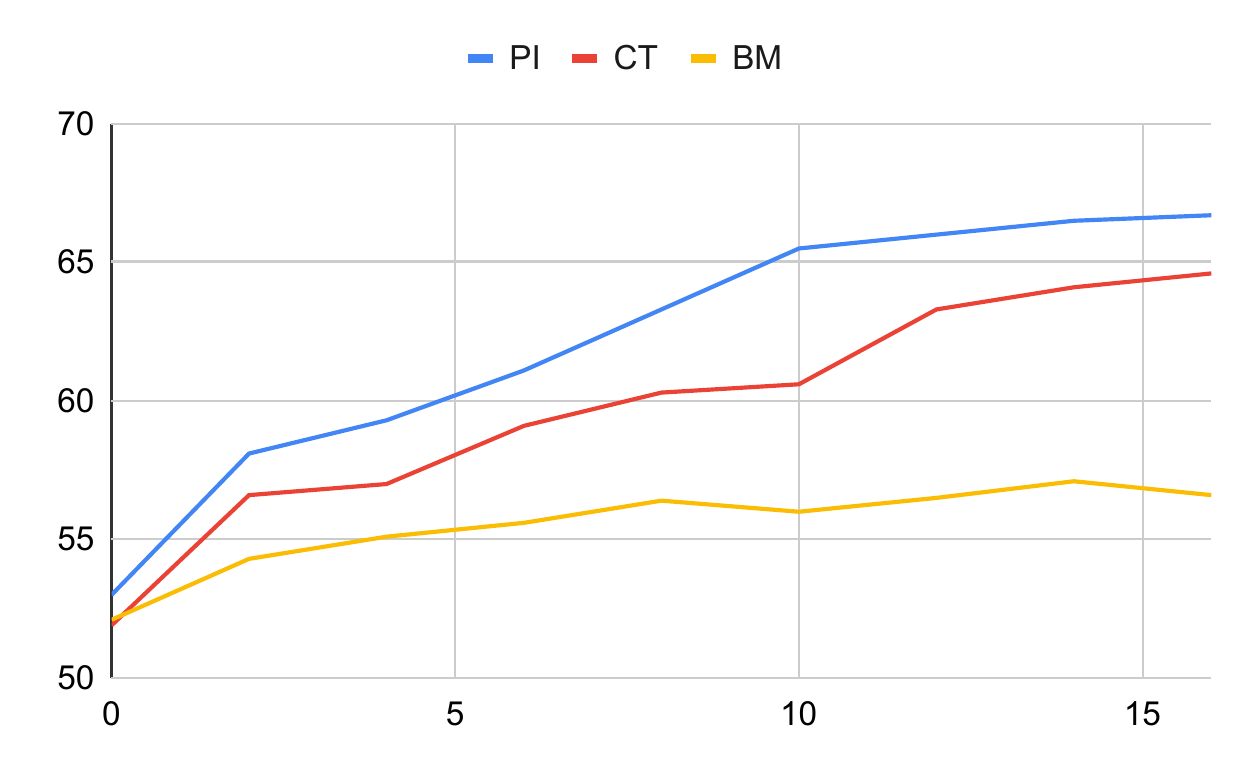}
  \caption{The Number of examples in few-shot learning for self-detection and detection performance. We average the results across all models for each leakage target.}
  \label{fig:few_shot}
\end{figure}

Finally, we investigate the impact of the number of examples used for few-shot learning in self-detection on the detection performance.
To do this, we compare the detection performance when varying the number of examples used for few-shot learning for each model.
\autoref{fig:few_shot} shows the detection performance when using different numbers of examples for few-shot learning.
The detection performance of self-detection improves as the number of examples increases.
On the other hand, when no examples are used, the performance significantly drops.
These results indicate that when using LLMs for detection, it is important to explicitly learn from leaked and non-leaked data examples.

\section{Ethical Considerations}
\label{sec:ethics}

In this paper, we conducted experiments using datasets containing sensitive information that needs to be protected, such as personal information and copyrighted works.
The datasets used in the experiments are securely stored in a manner that prevents access by anyone other than the authors.
We do not plan to publicly release these datasets.
Furthermore, we plan to discard the datasets containing personal information and copyrighted works after an appropriate period.
We used OpenAI's API, but since OpenAI does not use data inputted to their API for training, there is no concern about leakage.


\section{Conclusion}

Our comprehensive study on data leakage within LLMs brings to light several critical insights and innovations in mitigating the risks associated with the training of these models on massive web-crawled corpora.
Throughout our investigation, we have identified that even the minimal presence of personal information, copyrighted texts, and benchmark datasets in training data can lead to significant leakage, underscoring the imperative need for robust detection and prevention mechanisms.

By establishing and examining three key criteria proportion of leaked data, ease of generating such data, and detection rates of trained versus untrained data—we have delineated the nuanced relationship between the leakage rate and its implications on the output and detection rate.
This relationship is pivotal for understanding how even slight oversights in data handling can compromise the trustworthiness and integrity of LLMs.
Our experimental survey shed light on the alarming ease with which LLMs can generate sensitive and copyrighted material, challenging the assumption that the prevalence of such data in training sets is the sole determinant of leakage.

Our proposed self-detection simply employs few-shot learning.
It is known that techniques like chain-of-thought and discussion can improve performance when applied to inference~\cite{Wei2022ChainOT,Wang2022SelfConsistencyIC,Kaneko2023SolvingNP,Loem2023SAIEFS,Kaneko2024EvaluatingGB}. 
Therefore, it is conceivable that the performance of self-detection could be improved by using such techniques.

\bibliographystyle{named}
\bibliography{ijcai24}

\end{document}